\documentclass[sigconf]{acmart}

\AtBeginDocument{%
  }

\usepackage{pifont}

\usepackage{amssymb}
\usepackage{multirow}
\usepackage{amsmath}
\usepackage{algorithm}
\usepackage{algpseudocode}

\usepackage{booktabs}
\usepackage{makecell}
\usepackage{xcolor}
\usepackage{lipsum}

\setcopyright{acmlicensed}
\copyrightyear{2018}
\acmYear{2018}
\acmDOI{XXXXXXX.XXXXXXX}
\acmConference[Conference acronym 'XX]{Make sure to enter the correct
  conference title from your rights confirmation email}{June 03--05,
  2018}{Woodstock, NY}
\acmISBN{978-1-4503-XXXX-X/2018/06}

\begin{document}

% \title{AdvLoRA: Adversarial Low-Rank Adaptation of Vision-Language Models}

\title{Enhancing Adversarial Robustness of Vision-Language Models through Low-Rank Adaptation}

\author{Yuheng Ji$^{*}$}
\affiliation{%
  \institution{Institute of Automation, CAS\\
  School of Artificial Intelligence, UCAS}
  \country{}
  }

\author{Yue Liu$^{*}$}
\affiliation{%
  \institution{Institute of Data Science, \\ National University of Singapore}
  \country{}}

\author{Zhicheng Zhang}
\affiliation{%
  \institution{Institute of Automation, CAS\\
  School of Artificial Intelligence, UCAS}
  \country{}}

\author{Zhao Zhang}
\affiliation{%
  \institution{Beijing University of Posts and Telecommunications}
  \country{}}

\author{Yuting Zhao}
\affiliation{%
  \institution{Institute of Automation, CAS\\
  School of Artificial Intelligence, UCAS}
  \country{}}

\author{Xiaoshuai Hao}
\affiliation{%
  \institution{Beijing Academy of Artificial Intelligence}
  \country{}}

\author{Gang Zhou}
\affiliation{%
  \institution{Beijing University of Posts and Telecommunications}
  \country{}}

\author{Xingwei Zhang}
\affiliation{%
  \institution{Institute of Automation, CAS}
  \country{}}

% \vspace{-10pt}
\author{Xiaolong Zheng$^{\dagger}$}
\affiliation{%
  \institution{Institute of Automation, CAS\\
  School of Artificial Intelligence, UCAS}
  \country{}}

\renewcommand{\shortauthors}{Ji et al.}

\begin{abstract}
Vision-Language Models (VLMs) play a crucial role in the advancement of Artificial General Intelligence (AGI). As AGI rapidly evolves, addressing security concerns has emerged as one of the most significant challenges for VLMs.
In this paper, we present extensive experiments that expose the vulnerabilities of conventional adaptation methods for VLMs, highlighting significant security risks.
Moreover, as VLMs grow in size, the application of traditional adversarial adaptation techniques incurs substantial computational costs.
To address these issues, we propose a parameter-efficient adversarial adaptation method called \textbf{\textit{AdvLoRA}}  based on Low-Rank Adaptation.
We investigate and reveal the inherent low-rank properties involved in adversarial adaptation for VLMs.
Different from LoRA, we enhance the efficiency and robustness of adversarial adaptation by introducing a novel reparameterization method that leverages parameter clustering and alignment.
Additionally, we propose an adaptive parameter update strategy to further bolster robustness.
These innovations enable our AdvLoRA to mitigate issues related to model security and resource wastage. 
Extensive experiments confirm the effectiveness and efficiency of AdvLoRA.

\let\thefootnote\relax\footnotetext{$^{*}$ Equal Contribution.}
\let\thefootnote\relax\footnotetext{$^{\dagger}$ Corresponding Author.}
% \footnotetext{$^{*}$ Equal Contribution.}
% \footnotetext{$^{\dagger}$ Corresponding Author.}

% \blfootnote{$^{*}$ Equal Contribution.}
% \blfootnote{$^{\dagger}$ Corresponding Author.}

\end{abstract}

\begin{CCSXML}
<ccs2012>
   <concept>
       <concept_id>10010147.10010178</concept_id>
       <concept_desc>Computing methodologies~Artificial intelligence</concept_desc>
       <concept_significance>500</concept_significance>
       </concept>
 </ccs2012>
\end{CCSXML}

\ccsdesc[500]{Computing methodologies~Artificial intelligence}

\keywords{Adversarial robustness, Low-Rank adaptation, Vision-Language models}


\maketitle

\section{Introduction}

Artificial General Intelligence (AGI) aims to create intelligent agents that can perform as well as or better than humans on various cognitive tasks, making it a promising topic for research and industrial applications~\cite{pei2019towards, goertzel2014artificial}. As vision and language are key components of intelligence, Vision-Language Models (VLMs) have emerged as a crucial technique for achieving AGI~\cite{fei2022towards, achiam2023gpt}.

\begin{figure}[!t]
\centering
\small
\begin{minipage}{0.49\linewidth}
\centerline{\includegraphics[width=1\textwidth]{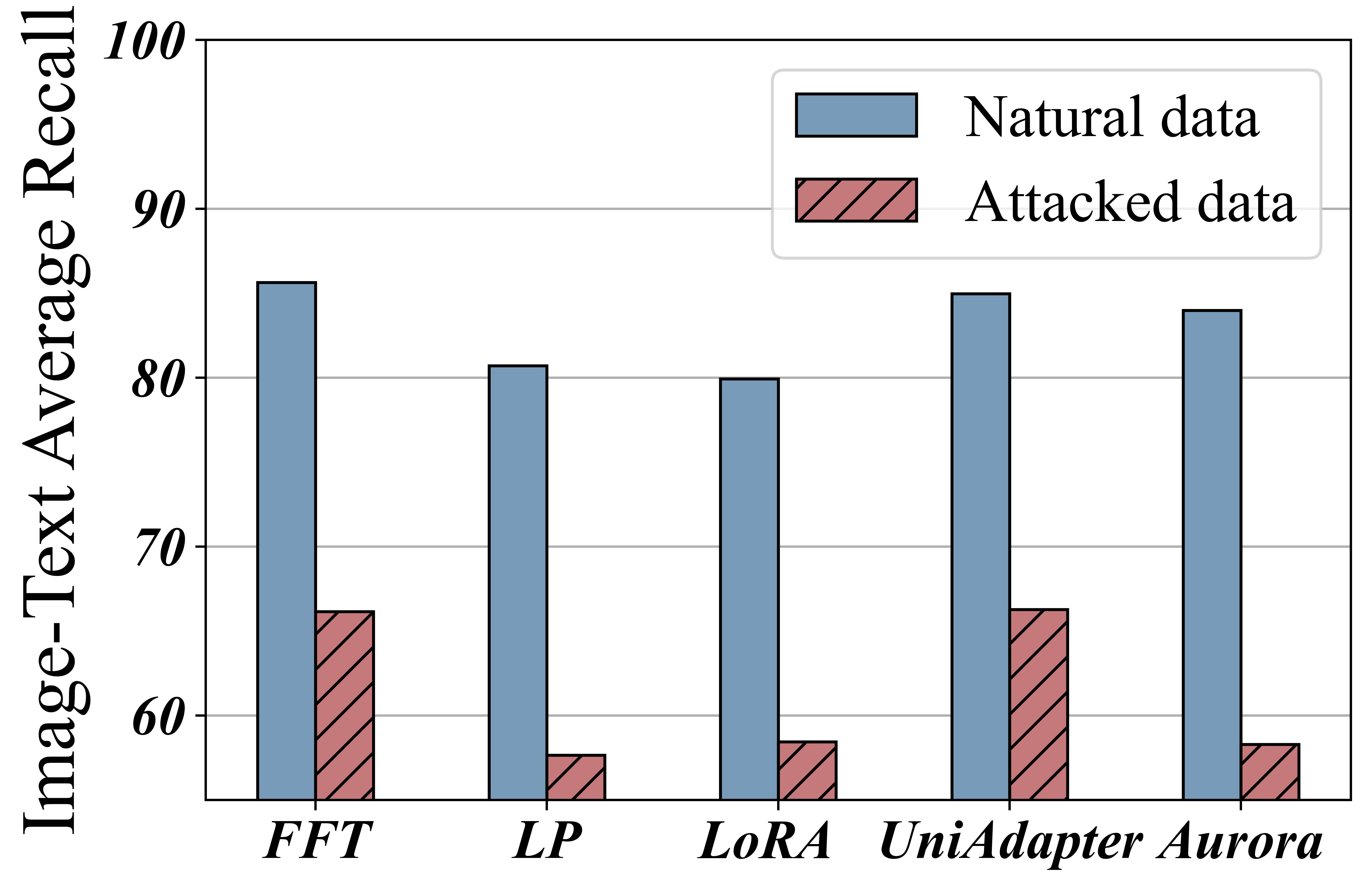}}
\centerline{(a) MSCOCO}
\vspace{3pt}
\end{minipage}
\begin{minipage}{0.49\linewidth}
\centerline{\includegraphics[width=1\textwidth]{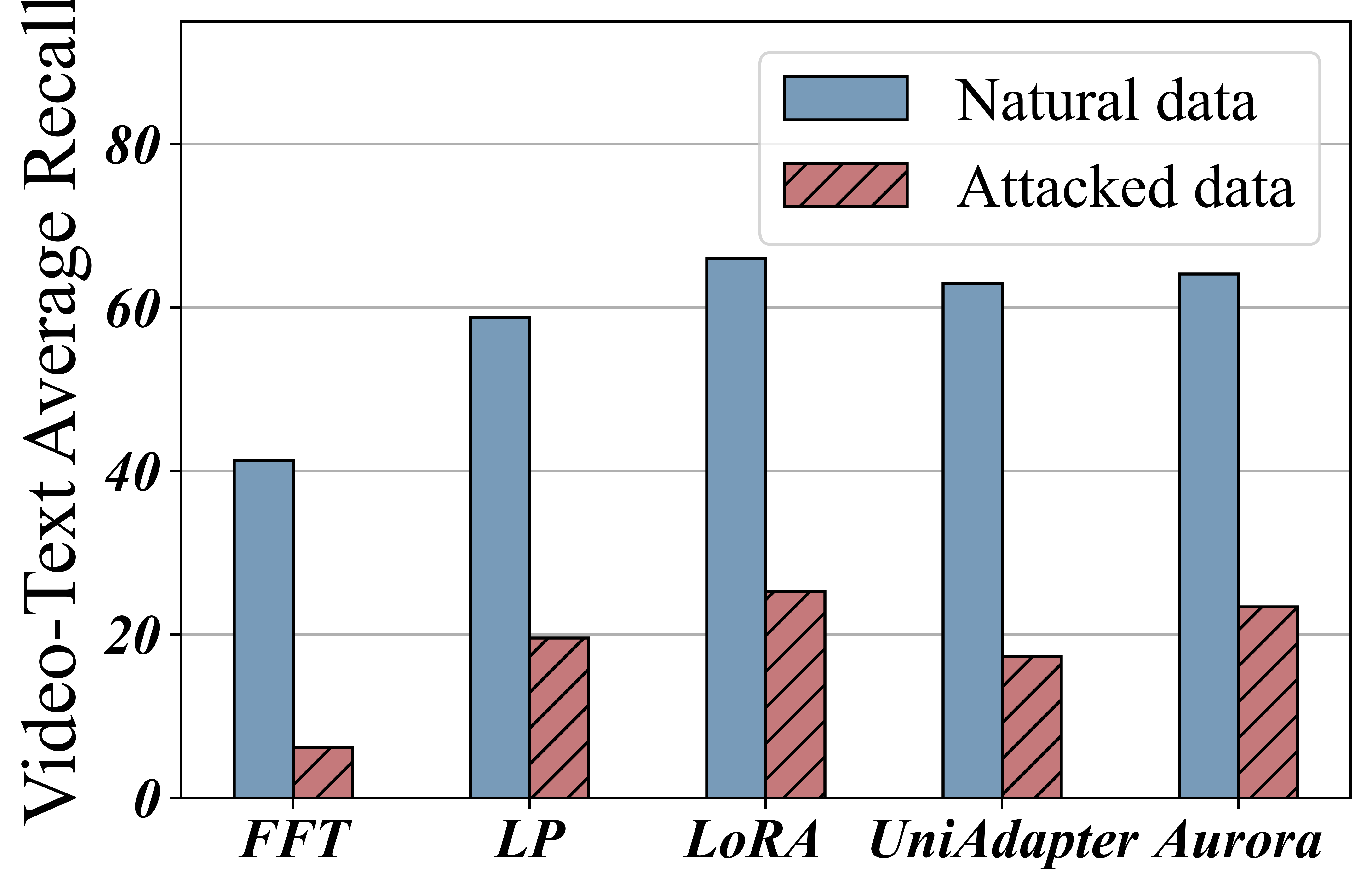}}
\centerline{(b) MSR-VTT}
\vspace{3pt}
\end{minipage}
\caption{Vulnerability of vision-language model adaptation methods to natural and adversarial data in MSCOCO (image-text data)~\cite{lin2014microsoft} and MSR-VTT (video-text data)~\cite{xu2016msr} datasets.}
\label{cuiruo}
\vspace{-1em}
\end{figure}

Recently, the adaptation of VLMs aims to improve the performance on different downstream tasks and has become a hot research topic. However, through extensive experiments, we find the vulnerability of the conventional adaptation methods, e.g., Full Fine-Tuning (FFT)~\cite{full-tuning-1,full-tuning-2,full-tuning-3}, Linear Probe (LP), LoRA~\cite{hu2021lora}, Unidapter~\cite{lu2023uniadapter}, and Aurora~\cite{Aurora} for VLMs, which may bring significant security threats in various domains, such as facial recognition~\cite{venkatesaramani2021re, sharif2016accessorize}, medical analysis~\cite{finlayson2019adversarial, ma2021understanding} and autonomous driving~\cite{zhang2022adversarial, feng2021intelligent}. As shown in Fig.~\ref{cuiruo}, we conduct adaptation experiments of VLMs on the natural and attacked data of the MSCOCO~\cite{lin2014microsoft} and MSR-VTT~\cite{xu2016msr} datasets. From these experimental results, we find that the average performance drops about 30.98\% on the attacked data. To solve this problem, various techniques are proposed against adversarial attacks by data augmentation~\cite{volpi2018generalizing, morris2020textattack}, attack detection~\cite{metzen2017detecting, liu2018adversarial} and adversarial training~\cite{goodfellow2014explaining, liu2020adversarial}. As the most effective defense strategy, adversarial training enhances the adversarial robustness of VLMs by retraining the model on mined adversarial examples~\cite{madry2017towards, szegedy2013intriguing, pang2020bag}.

\begin{figure}[!t]
    \centering
    \includegraphics[width=\linewidth]{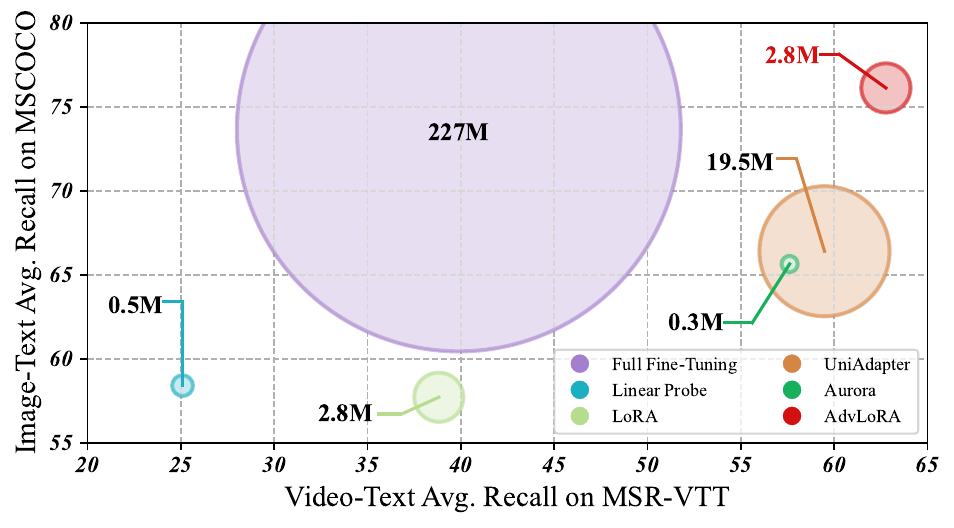}
    \caption{Adversarial robustness and tunable parameter number of adversarial adaptation methods on two dataset.}
    \label{bubble}
    \vspace{-1em}
\end{figure}

However, as the sizes of VLMs increase, the conventional adversarial training method with full parameter updating to improve the adversarial robustness of VLMs will lead to high computing and storage costs~\cite{villa}. In recent years, Parameter-Efficient Fine-Tuning (PEFT) technology has garnered widespread attention as a novel adaptation paradigm due to its significant success in adapting large-scale pre-trained models. PEFT technologies can adapt VLMs with extremely small additional tunable parameters and achieve comparable or better performance than FFT methods. While PEFT technologies have demonstrated remarkable success in natural scenarios, their application in adversarial attack scenarios remains largely uncharted territory. But simply applying the adversarial training on the conventional adaptation methods will lead to 1) limited defense performance and 2) high computational and storage costs. To verify our points, we visualize the adversarial robustness performance and the tunable parameter number of different adversarial adaptation methods in Fig.~\ref{bubble}. From the results, we find that the existing adaptation methods such as FFT and UniAdapter will lead to large parameter costs. Besides, LoRA, LP, and Aurora are not robust to adversarial attacks.

To solve these problems, we aim to develop a parameter-efficient adversarial adaptation method termed AdvLoRA to effectively and efficiently improve the robustness of VLMs against attacks. At first, similar to LoRA, the intrinsic low-rank property of adversarial adaptation for VLMs is revealed. Secondly, we improve LoRA with a novel reparameterizing technology. Concretely, we regard the rank of LoRA as the number of cluster centers and utilize the clustering algorithm to reparameterize LoRA from the weight matrices of VLMs. The weight matrices are decoupled into the clustering centers and the clustering distribution matrices. Subsequently, we impose constraints on their product to align with the parameter distribution of the original weight matrix. Moreover, we design an adaptive parameter update strategy to improve the robustness further. Through these settings, we effectively and efficiently facilitate the adversarial adaptation of VLMs. Our designs on low-rank for adversarial adaptation are motivated by the common dense direction theory~\cite{allen2022feature}, which demonstrates that low-rank adaptation in shallow convolutional neural networks are more suitable to effectively enhance their robustness. For the first time, this paper empirically verifies the applicability of this theory to VLMs and introduces a novel clustering-based initialization method for LoRA, facilitating the process of adversarial fine-tuning. The contributions of this paper are summarized as follows:

\begin{itemize}

\item We demonstrate the vulnerability of VLMs with different adaptation methods to adversarial attacks via experiments.

\item We investigate and reveal the intrinsic low-rank property during the adversarial adaptation for VLMs. 

\item We propose a novel parameter-efficient adversarial adaptation method named AdvLoRA with parameter clustering, parameter alignment, and adaptive parameter update.

\item We are the first to introduce the adversarial adaptation for VLMs. Extensive experiments demonstrate the effectiveness and efficiency of our proposed method. 

\end{itemize}

\section{Related Work}
\textbf{Parameter-efficient Tuning on VLMs.}
Vision-Language Models (VLMs) excel in various vision-language tasks, such as cross-modal retrieval~\cite{hao2023dual,hao2023mixgen,hao2024uncertainty,hao2021multi,hao2022listen,hao2021matters} and generation~\cite{bao2023one,rombach2022high}. However, they may struggle when task data diverges from training data, necessitating re-training or fine-tuning on task-specific datasets. Traditional adaptation methods like Full Fine-Tuning (FFT) become inefficient with larger VLMs, prompting the need for parameter-efficient tuning to reduce training and storage costs.
Recent approaches inspired by natural language processing~\cite{ liu2023gpt, zhang2023adaptive, dettmers2023qlora} and computer vision~\cite{ jia2022visual, bahng2022exploring} aim to adapt frozen VLMs using minimal tunable parameters, achieving results comparable to full parameter tuning. These methods fall into three categories: adapter-based~\cite{lu2023uniadapter,gao2023clip}, prompt-based~\cite{zhou2022learning,clip_prompt1,clip_prompt2}, and LoRA-based~\cite{zhong2024multi, qiang2024bilora, liu2024aflora, wang2024lora, zhao2024galore, pan2024lisa}. LoRA-based approaches are particularly notable for their reduced tunable parameters, lack of additional input, and minimal inference latency. This paper addresses suboptimal initialization in standard LoRA methods and explores a clustering-based reparameterization strategy to enhance VLM robustness during adaptation.

\textbf{Adversarial Adaptation on VLMs.}
Some researchers have shown that artificial neural networks, including Vision-Language Models (VLMs), are vulnerable to unrecognized attacks~\cite{ cai2023clap, zhao2023evaluating, liu2025guardreasoner, liu2024flipattack}. Specifically, adding perturbations to inputs can lead VLMs to make incorrect decisions with high confidence. To enhance adversarial robustness, many studies focus on data augmentation~\cite{cai2023clap,wise-ft} and adversarial training~\cite{villa,zsrobust,diao2024tasar}. Adversarial training, particularly effective, improves robustness by incorporating adversarial inputs into the training process via a min-max formulation~\cite{madry2017towards}.
Initially, some efforts used adversarial training techniques to train VLMs from scratch~\cite{villa}. Recently, adversarial adaptation has emerged as a cost-effective strategy for enhancing adversarial robustness post-pretraining~\cite{ xuautolora, zsrobust, yuan2024fulllora, zhang2023adversarial}. However, most methods update all parameters of the pre-trained model and focus on visual models. A few multi-modal approaches, such as TeCoA~\cite{zsrobust}, utilize prompt tuning for adversarial adaptation.

In this paper, we introduce a parameter-efficient LoRA method for adversarial adaptation in cross-modal tasks. Unlike AutoLoRa~\cite{xuautolora}, which updates all parameters while addressing gradient instability, our approach maintains efficiency. Inspired by dense direction theory~\cite{allen2022feature}, which suggests that low-rank adaptation enhances robustness in shallow networks, we provide the first empirical validation of this theory in VLMs and present a novel clustering-based initialization method for LoRA to streamline adversarial fine-tuning.

\section{Method}
In Sec.~\ref{task_definition}, we first define the cross-modal retrieval. Subsequently, addressing the vulnerability of VLMs to adversarial attacks, we introduce an adversarial training module in Sec.~\ref{adv_train_module} to enhance the model's adversarial robustness. Finally, to mitigate the high cost associated with adversarial training, we present an adaptation module in Sec.~\ref{adaptation_module}, which maintains the VLMs' adversarial robustness while reducing the expenses of adversarial training.

\subsection{Task Definition}
\label{task_definition}
\textbf{Cross-Modal Retrieval.}
Cross-modal retrieval aims to utilize information from one modality to retrieve semantically relevant information from another. We select cross-modal retrieval as our benchmark task due to its efficacy in assessing the quality of cross-modal representation learning in VLMs. Under adversarial attacks, cross-modal retrieval serves as an effective metric for evaluating whether models can learn robust feature representations.

Taking image-to-text retrieval as an example, given an image $v_i$, its semantic representation $\boldsymbol{z}_i^v=\mathcal{F}_{v}(v_i)$ is used to compute the cosine similarity with each textual representation $\boldsymbol{z}_j^w$ within the text database as follows:
\begin{equation}
\text{sim}(\boldsymbol{z}_i^v, \boldsymbol{z}_j^w)=\frac{\boldsymbol{z}_i^v \cdot \boldsymbol{z}_j^w}{\|\boldsymbol{z}_i^v\| \|\boldsymbol{z}_j^w\|},
\label{similarity}
\end{equation}
where $\boldsymbol{z}_j^w=\mathcal{F}_{w}({w_j})$ represents the semantic representation derived from the textual data $w_j$ after feature extraction via the text encoder $\mathcal{F}_{w}$. Then we select the highest similarity text data as the retrieval results. Under adversarial attacked, robust VLMs could learn semantically invariant feature representations so that they will not be misled by small perturbations.

\subsection{Adversarial Training Module}
\label{adv_train_module}
Extensive experimentation demonstrated that both VLMs and their variants adapted with PEFT methods are susceptible to adversarial attacks, as illustrated in Fig.~\ref{cuiruo}. Consequently, in this subsection, we design an adversarial training module to enhance the adversarial robustness of VLMs. We begin by introducing the concept of adversarial attacks, followed by the presentation of adversarial training as an effective defense technology for enhancing adversarial robustness.

\textbf{Adversarial Attack.}
Adversarial attacks $\delta$ is a tensor added to the natural image $v$, $v_a=v+\delta$, aiming to fool the model into making the incorrect decision as formulated:

\begin{equation}
v_a=\mathop{\arg \max}_{v_a}\mathcal L(v_a,w), \quad \text{s.t.} \quad \Vert{v_a-v}\Vert_{p} \leq \varepsilon,
\label{attack}
\end{equation}
where $p$ donates the $p$-norm, and $\varepsilon$ donates the restriction value of values, which is often set to be smaller than $8/255$. Thus, the adversarial attacks are imperceptible to humans. In this paper, we focus on adversarial attacks on visual data, as attacks on natural language are readily perceptible to humans. Therefore, it is practically significant and more challenging to make attacks on visual data. Concretely, we utilize PGD~\cite{madry2017towards} to generate $v_a$ as follows:
\begin{equation}
v_a = \prod\left\{\text{clip}_{\varepsilon}( v + \xi \cdot \text{sign}\left(\nabla_v \mathcal L(v,w)\right))\right\},
\label{pgd}
\end{equation}
where $\text{sign}(\nabla_v \mathcal L(v,w)))$ denotes the sign value of the back-propagated gradient. Besides, $\xi$ is the step size of each iteration. And $\text{clip}_{\varepsilon}(x)=\min(x,\varepsilon)$ clips each value of $x$ to be smaller than $\varepsilon$ and return $\varepsilon$ when the value of any dimension exceeds $\varepsilon$. $\prod\{\cdot\}$ denotes the iterative procedure. In this manner, $v_a$ can fool the model to make the incorrect decision. Notably, for video data, we treat it as a collection of images and attack 20\% of the frames by randomly sparse sampling~\cite{wei2019sparse}.

\textbf{Adversarial Training.}
Adversarial training technologies refer to retraining the model on attacked data, which can learn semantically invariant features under adversarial attacks. Adversarial training aims to minimize the following objective:
\begin{equation}
\theta=\mathop{\arg \min}_{\theta}\mathcal L(v_a,w),
\label{at}
\end{equation}
where $\theta$ donates the parameters of the model.

\subsection{Adaptation Module}
\label{adaptation_module}
Although adversarial training can effectively enhance VLMs' adversarial robustness, it requires updating all parameters based on gradient information, leading to a significant cost overhead. To alleviate this issue, in this subsection, we propose an adaptation module that performs adversarial training on LoRA to reduce the number of tunable parameters, achieving parameter-efficient adversarial adaptation. We first provide a brief introduction to LoRA, followed by the introduction of clustering reparameterization and parameter alignment methods, as well as an adaptive parameter update strategy, to facilitate adversarial adaptation.

\textbf{LoRA.}
LoRA achieves parameter-efficient adaptation by updating two low-rank matrices attached to the frozen pre-trained weights. Specifically, given the pre-trained weights $\boldsymbol{W_0} \in \mathbb{R}^{m \times n}$, and the LoRA matrices $\boldsymbol{A} \in \mathbb{R}^{m \times k}$, $\boldsymbol{B} \in \mathbb{R}^{k \times n}$, the input $\boldsymbol{X}^{(l-1)} \in \mathbb{R}^{b \times m}$ is processed through the following computation to obtain the output $\boldsymbol{X}^{(l)} \in \mathbb{R}^{b \times n}$ as follows:
\begin{equation}
\boldsymbol{X}^{(l)} = \boldsymbol{X}^{(l-1)}\boldsymbol{W_0} + \boldsymbol{X}^{(l-1)}\boldsymbol{AB},
\label{lora}
\end{equation}
where $k \ll \min(m, n)$. And $\boldsymbol{A}$ and $\boldsymbol{B}$ are initialized as follows:
\begin{equation}
\boldsymbol{A} \sim \mathcal{N}(0, \sigma^2), \quad \boldsymbol{B} = \textbf{0},
\label{init}
\end{equation}
where $\mathcal{N}$ denotes the Gaussian distribution. 

During the adaptation process, $\boldsymbol{W_0}$ is fixed, while $\boldsymbol{A}$ and $\boldsymbol{B}$ are updated via the gradient descent methods. In our proposed model, AdvLoRA, we freeze $\boldsymbol{W_0}$ and solely update $\boldsymbol{A},\boldsymbol{B}$ through adversarial adaptation to achieve adversarial robustness in the model as follows:

\begin{equation}
\theta_{\boldsymbol{A},\boldsymbol{B}}=\mathop{\arg \min}_{\theta_{\boldsymbol{A},\boldsymbol{B}}}\mathcal L(v_a,w).
\label{update}
\end{equation}
Our model adheres to conventional practice by incorporating LoRA into both the attention modules and feed-forward networks in BLIP.

\textbf{Reparameterization and Adaptive Parameter Update.}
The primary distinction between AdvLoRA and other LoRA-like methods lies in the parameterization process of the matrices $\boldsymbol{A}, \boldsymbol{B}$. In the original LoRA, a random Gaussian initialization for $\boldsymbol{A}$ and zero for $\boldsymbol{B}$, so $\boldsymbol{AB}$ is zero at the beginning of adaptation. In contrast, our model, AdvLoRA, initially performs clustering on the weight matrix $\boldsymbol{W_0}$ of the pre-trained model, treating the rank $k$ of LoRA as the number of cluster centers. Specifically, given an weight matrix $\boldsymbol{W} \in \mathbb{R}^{m \times n}$ and the rank $k$, we first randomly initialize $k$ cluster center $ \boldsymbol{C}=\{\boldsymbol{c_1}, \boldsymbol{c_2}, \ldots, \boldsymbol{c_k}\}$. Then, for each column $\boldsymbol{w_i}$ of $\boldsymbol{W}$, compute the distances to each cluster center $\boldsymbol{c_j}$ and assign $\boldsymbol{w_i}$ to the closest cluster as follows:
\begin{equation}
\textrm{cluster}_i=\mathop{\arg \min}_{j} \Vert{\boldsymbol{w_i}-\boldsymbol{c_j}}\Vert_{2}.
\label{cluster_distance}
\end{equation}
Then update the cluster centers by computing the mean of all data points assigned to each cluster as follows:
\begin{equation}
\boldsymbol{c_j} = \frac{1}{|\boldsymbol{S}_j|} \sum_{\boldsymbol{w_i} \in \boldsymbol{S}_j} \boldsymbol{w_i},
\label{cluster_center}
\end{equation}
where \( \boldsymbol{S}_j \) is the set of columns of $\boldsymbol{W}$ assigned to cluster \( j \).
Repeat the above steps until the cluster centers no longer change significantly or a maximum number of iterations is reached. In this manner, we obtain the cluster center embeddings $\boldsymbol{C} \in \mathbb{R}^{k \times n}$ and the distance assignment matrix $\boldsymbol{D} \in \mathbb{R}^{m \times k}$, where each element \( d_{ij} \) represents the distance between the $\boldsymbol{w}_i$ and cluster center $\boldsymbol{c}_j$. The distance assignment matrix \( \boldsymbol{D} \) can be computed using the following formula:
\begin{equation}
 \boldsymbol{d}_{ij} = \|\boldsymbol{w_i} - \boldsymbol{c_j}\|_2.
\label{a}
\end{equation} And the cluster center representation matrix \( \boldsymbol{C} \) is simply the matrix of cluster centers as follows:
\begin{equation}
 \boldsymbol{C} = \begin{bmatrix} \boldsymbol{c_1} , \boldsymbol{c_2} , \ldots , \boldsymbol{c_k} \end{bmatrix}.
\label{b}
\end{equation}
After the parameter clustering, the clustering assignment matrix $\boldsymbol{D} \in \mathbb{R}^{m \times k}$ and the parameter center $\boldsymbol{C} \in \mathbb{R}^{k \times n}$ can be represented the $\boldsymbol{A} \in \mathbb{R}^{m \times k}$ and $\boldsymbol{B} \in \mathbb{R}^{k \times n}$ in the original LoRA method. By these settings, we provide a better reparameterization of the tunable parameters in LoRA. It separates the parameters into different clusters, which have different functions in the whole network. 

After obtaining the matrices $\boldsymbol{A}$ and $\boldsymbol{B}$, we further impose constraints on their product $\boldsymbol{AB}$ to align with the parameter distribution of the original weight matrix $\boldsymbol{W}_0$ as follows:

\begin{equation}
\min{\Vert{\boldsymbol{W_0}-\boldsymbol{AB}}\Vert_{2}}.
\label{align}
\end{equation}
In this manner, we can ensure the initialization of $\boldsymbol{AB}$ is close to $\boldsymbol{W_0}$ at the beginning of the training.

During the process of model adversarial adaptation, we design an adaptive update parameter, $\alpha$, to facilitate the model's adaptive learning of robust semantic representations as follows:

\begin{equation}
\boldsymbol{Y} = \boldsymbol{XW_0} + \alpha \cdot \boldsymbol{XAB}.
\label{balance}
\end{equation}
$\alpha$ is a tunable neural network parameter, which can control the adaptation rate during the adversarial adaptation. In summary, we delineate the entire workflow of AdvLoRA in Algorithm \ref{algorithm_1}.

% ALGORITHM
\begin{algorithm}[t]
\small
\caption{AdvLoRA WorkFlow on VLMs.}
\label{algorithm_1}
\begin{algorithmic}[1] % [1] adds line numbers
    \Require % Input
    Images: $\boldsymbol{V}=\{{v}_1, {v}_2, \ldots, {v}_n\}$; 
    Texts: $\boldsymbol{W}=\{{w}_1, {w}_2, \ldots, {w}_n\}$; 
    Visual encoder: $\mathcal{F}_v$; 
    Textual encoder: $\mathcal{F}_w$; 
    Pre-trained weight matrix: $\boldsymbol{W}_0$; 
    LoRA matrix: $\boldsymbol{A,B}$; 
    Adaptive parameter: $\alpha$; 
    Restriction value: $\epsilon$; 
    PGD step: $\xi$; 
    Loss function: $\mathcal L$.
    
    \Ensure % Output
    Representations of $\boldsymbol{V}$ and $\boldsymbol{W}$: 
    $\boldsymbol{Z^v}=\{\boldsymbol{z}_1^v, \boldsymbol{z}_2^v, \ldots, \boldsymbol{z}_n^v\}$, 
    $\boldsymbol{Z^w}=\{\boldsymbol{z}_1^w, \boldsymbol{z}_2^w, \ldots, \boldsymbol{z}_n^w\}$.

    \While{at adversarial fine-tuning stage}
        \State Perform clustering algorithm on $\boldsymbol{W_0}$ and obtain cluster center representation in Eq. \eqref{cluster_distance} and Eq. \eqref{cluster_center};
        \State Obtain the LoRA matrix $\boldsymbol{A, B}$ from cluster center representation and $W_0$ in Eq. \eqref{a} and Eq. \eqref{b};
        \State Impose constraints on $\boldsymbol{A}$ and $\boldsymbol{B}$ with SGD algorithm in Eq. \eqref{align};
        \State Calculate the loss $l$ using $\boldsymbol{V}$, $\boldsymbol{W}$, $\boldsymbol{Y}$, $\mathcal{F}_v$, $\mathcal{F}_w$, and the loss function $\mathcal L$ in Eq. \eqref{lora}.
        \State Obtain the adversarial attack $\delta$ with $\boldsymbol{V}$, $\epsilon$, $k$, $\xi$ and $l$ in Eq. \eqref{pgd}.
        \State Add $\delta$ to original images $\boldsymbol{V}$ to obtain the attacked images $\boldsymbol{V_a}$.
        \State Update $\boldsymbol{A}, \boldsymbol{B}$ via Eq. \eqref{at}.
    \EndWhile
    \State Generate robust representations of $\boldsymbol{V}$ and $\boldsymbol{W}$ to downstream tasks with adversarially adapted $\boldsymbol{A,B}$ and $\mathcal{F}_v$, $\mathcal{F}_w$.
\end{algorithmic}
\end{algorithm}

\section{Experiment}
\subsection{Experimental Setup}
\label{datasets}
\textbf{Datasets.}
We conducted a comprehensive evaluation of our proposed model, AdvLoRA, across two types of retrieval tasks and four widely used datasets. This evaluation highlights AdvLoRA's superior performance in cross-modal understanding tasks, specifically image-text retrieval with Flickr30K~\cite{plummer2015flickr30k} and MSCOCO~\cite{lin2014microsoft}, as well as video-text retrieval with DiDeMo~\cite{anne2017localizing} and MSRVTT~\cite{xu2016msr}.

\begin{itemize}
\item{\textbf{Flickr30K}}~\cite{plummer2015flickr30k} contains 31,783 images and 158,915 captions totally. Each image is often annotated with 5 captions. Following the split in Uniadapter~\cite{lu2023uniadapter} and Aurora~\cite{Aurora}, we use 1,000 images for testing, another 1,000 for validation, and the rest for training.

\item{\textbf{MSCOCO}}~\cite{lin2014microsoft} is a large dataset containing 123,287 images and 616,435 captions. Each image is annotated with 5 captions. Following the split in Uniadapter~\cite{lu2023uniadapter} and Aurora~\cite{Aurora}, we use 5,000 images for testing, another 5,000 for validation, and the rest for training.

\item{\textbf{Didemo}}~\cite{anne2017localizing} contains 10,000 videos and 40,000 annotations. Following Frozen in Time~\cite{bain2021frozen}, we concatenate all descriptions corresponding to the same video into a single sentence to conduct actually video-paragraphto retrieval task.

\item{\textbf{MSR-VTT}}~\cite{xu2016msr} is a popular video-text dataset. It contains 10,000 video and 200,000 captions. Following the split in Uniadapter~\cite{lu2023uniadapter} and Aurora~\cite{Aurora}, we use 1,000 videos for testing, another 9,000 for training.
\end{itemize}

\label{baselines}
\textbf{Baselines.}
We compare AdvLoRA with conventional adaptation methods, which are implemented by BLIP: full fine-tuning (BLIP-FFT), linear probe (BLIP-LP); as well as the PEFT method on BLIP: LoRA(BLIP-LoRA), Aurora, and Uniadapter. 
\begin{itemize}
\item{\textbf{BLIP-FFT}}~\cite{blip} is a conventional adaptation technique that enhances the performance of BLIP for specific downstream tasks by retraining and updating full parameters in downstream tasks.

\item{\textbf{BLIP-LP}}~\cite{blip} is an adaptation technique that involves adding and training a linear layer on top of the frozen pre-trained model BLIP to adapt to specific tasks.

\item{\textbf{BLIP-LoRA}}~\cite{hu2021lora} is a PEFT technology that adapts BLIP by introducing low-rank adapters to capture task-specific information, allowing for efficient adaptation to downstream tasks with minimal tunable parameter updates.

\item{\textbf{Uniadapter}}~\cite{lu2023uniadapter} is the first adapter-based PEFT technology for parameter-efficient cross-modal adaptation.

\item{\textbf{Aurora}}~\cite{Aurora} is a parameter-efficient cross-modal transfer learning framework that uses mode approximation to generate a minimal set of tunable parameters, achieving lightweight multi-modal adaptation.

\end{itemize}

\textbf{Metrics.}
We employ $Recall@k$ as our evaluation metric, measuring the proportion of relevant items retrieved within the top $k$ results. A higher $Recall@k$ indicates better performance in retrieving relevant items, reflecting the model's effectiveness and reliability in information retrieval tasks.

\textbf{Implementations.}
\label{setting}
Our implementation is based on Salesforce's open-source codebase~\cite{blip}. Following~\cite{lu2023uniadapter, Aurora}, we also apply BLIP~\cite{blip} as our vision-language backbone for all tasks. We use PyTorch to implement all experiments on the NVIDIA V100 GPU (32GB). We employ PGD-3~\cite{madry2017towards} for adversarial adaptation and to assess the model's robustness. For the video-text retrieval task, we follow the work of Wei et al.~\cite{wei2019sparse} by adopting an attack strategy that sparsely samples 20\% of the video frames. Furthermore, we adopt the setup of BLIP, utilizing a momentum encoder to enhance the retrieval performance of our model. To ensure a fair comparison, the momentum encoder is also applied to the other baseline methods. During the fine-tuning process, the parameters of the backbone model are kept frozen. We present the hyperparameter setting in Tab.~\ref{hyper}.

\begin{table}[t]
    \caption{Hyperparameter setting}
    \small
    \centering
    \scalebox{0.88}{
    \begin{tabular}{c c c c c}
    \hline
        \multirow{2}{*}{config} & \multicolumn{2}{c}{Image-text Retrieval} & \multicolumn{2}{c}{Video-text Retrieval}  \\ 
        ~ & Flickr30K & MSCOCO & Didemo & MSR-VTT \\ \hline
        optimizer & AdamW & AdamW & AdamW & AdamW \\ 
        lr & 1e-5 & 1e-5 & 1e-4 & 1e-4 \\ 
        schedule & cosine decay & cosine decay & cosine decay & cosine decay \\ 
        training bs & 16 & 16 & 8 & 8 \\ 
        inference bs & 32 & 32 & 8 & 8 \\ 
        frames & - & - & 16 & 16 \\ 
        attack ratio & - & - & 20\% & 20\% \\ 
        epochs & 5 & 5 & 5 & 5 \\ 
        training input & 384 & 384 & 8*224 & 8*224 \\ 
        inference input & 384 & 384 & 16*224 & 16*224 \\ 
        attack type & PGD-3 & PGD-3 & PGD-3 & PGD-3 \\ 
        attack alpha & 1/255 & 1/255 & 1/255 & 1/255 \\
        PGD-epsilon & 1/255 & 1/255 & 1/255 & 1/255 \\ 
        rank & 10 & 10 & 10 & 10 \\ 
        adaptive weight & 1e-3 & 1e-3 & 1e-3 & 1e-3 \\ 
        weight norm lr & 1e-3 & 1e-3 & 1e-3 & 1e-3 \\ \hline
    \end{tabular}
    }
    \label{hyper}
\vspace{-1em}
\end{table}

\begin{table*}[!ht]
\caption{Adversarial experiment on MSCOCO. An asterisk (*) indicates that adversarial adaptation has been performed. The best results are displayed in bold, while the second-best results are underlined.}
    \centering
    \scalebox{0.88}{
    \begin{tabular}{lcccccccccc}
    \hline
        \multirow{2}{*}{Method} & \multirow{2}{*}{Tunable Para.} & \multicolumn{4}{c}{MSCOCO TR} & \multicolumn{4}{c}{MSCOCO IR} \\
        ~ & ~ & R@1 & R@5 & R@10 & R@Mean & R@1 & R@5 & R@10 & R@Mean & Mean \\ \hline
        BLIP+FFT+Att & 223M & 53.38 & 75.12 & 82.62 & 70.37 & 42.25 & 67.03 & 76.47 & 61.92 & 66.15 \\
        BLIP+FFT*+Att & 223M & \underline{65.42} & \underline{84.68} & \underline{89.4} & \underline{79.83} & \underline{47.62} & \underline{73.43} & \underline{81.35} & \underline{67.47} & \underline{73.65} \\
        BLIP+LoRA+Att & 2.8M & 43.2 & 66.2 & 74.8 & 61.4 & 35.85 & 60.4 & 70.16 & 55.47 & 58.44 \\
        BLIP+LoRA*+Att & 2.8M & 42.22 & 66.12 & 74.7 & 61.01 & 34.69 & 59.39 & 69.14 & 54.41 & 57.71 \\
        BLIP+LP+Att & 0.5M & 43.22 & 65.82 & 74.46 & 61.17 & 34.6 & 58.59 & 68.86 & 54.12 & 57.61 \\
        BLIP+LP*+Att & 0.5M & 44.14 & 67.18 & 76.04 & 62.45 & 34.57 & 59.14 & 69.3 & 54.34 & 58.40 \\
        UniAdapter+Att & 19.5M & 53.98 & 75.66 & 82.74 & 70.79 & 42.02 & 66.8 & 76.39 & 61.74 & 66.27 \\
        UniAdapter*+Att & 19.5M & 50.76 & 76.68 & 85.4 & 70.95 & 39.9 & 67.8 & 77.88 & 61.86 & 66.40 \\
        Aurora+Att & 0.3M & 44.56 & 67.04 & 75 & 62.2 & 34.98 & 59.34 & 68.75 & 54.36 & 58.28 \\
        Aurora*+Att & 0.3M & 54.56 & 77.68 & 84.52 & 72.25 & 40.08 & 60.17 & 75.66 & 60.17 & 65.64 \\
        AdvLoRA+Att & 2.8M& 46.76 & 69.18 & 76.72 & 64.22 & 37 & 61.25 & 70.76 & 56.34 & 60.28 \\
        AdvLoRA*+Att & 2.8M & \textbf{67.28} & \textbf{87.16} & \textbf{92.76} & \textbf{82.4} & \textbf{49.02} & \textbf{75.88} & \textbf{84.59} & \textbf{69.83} & \textbf{76.12} \\ \hline
    \end{tabular}
    }
    \label{advlora_coco_att}
\end{table*}

\begin{table*}[!ht]
    \caption{Adversarial experiment on Flickr30K. An asterisk (*) indicates that adversarial adaptation has been performed. The best results are displayed in bold, while the second-best results are underlined.}
    \centering
    \scalebox{0.88}{
    \begin{tabular}{lcccccccccc}
    \hline
        \multirow{2}{*}{Method} & \multirow{2}{*}{Tunable Para.} & \multicolumn{4}{c}{Flickr30K TR} & \multicolumn{4}{c}{Flickr30K IR} \\
        ~ & ~ & R@1 & R@5 & R@10 & R@Mean & R@1 & R@5 & R@10 & R@Mean & Mean \\ \hline
        BLIP+FFT+Att & 223M & 21.1 & 38.4 & 46 & 35.16 & 21.96 & 42.62 & 51.18 & 38.58 & 36.87 \\
        BLIP+FFT*+Att & 223M & 64.6 & {84.8} & {87.7} & 79.03 & {55.06} & {79.52} & {84.46} & \underline{73.01} & {76.02} \\
        BLIP+LoRA+Att & 2.8M & 67 & 81.8 & 84.2 & 77.67 & \underline{58.5} & 77.48 & 82.7 & 72.89 & 75.28 \\
        BLIP+LoRA*+Att & 2.8M & 65.6 & \textbf{87.1} & \underline{89.5} & 80.4 & 54.62 & 79.92 & 85.18 & 73.24 & 76.82 \\
        BLIP+LP+Att & 0.5M & 55.9 & 76 & 81.7 & 71.2 & 49.3 & 70.82 & 77.48 & 65.87 & 68.53 \\
        BLIP+LP*+Att & 0.5M & 56.1 & 75.7 & 82.7 & 71.5 & 48.14 & 70.5 & 78.18 & 65.61 & 68.55 \\
        UniAdapter+Att & 19.5M & 67.2 & 82.5 & 86.5 & 78.73 & 58.26 & 77.26 & 83.3 & 72.94 & 75.84 \\
        UniAdapter*+Att & 19.5M & \textbf{71.2} & {85.8} & 88.2 & \underline{81.73} & \textbf{59.12} & \textbf{80.4} & \underline{85.82} & \textbf{75.11} & \underline{78.42} \\
        Aurora+Att & 0.3M & 65.4 & 80.7 & 84.4 & 76.83 & 56.98 & 76.64 & 82.22 & 71.95 & 74.39 \\
        Aurora*+Att & 0.3M & {69.1} & 84.1 & 87.3 & {80.17} & 56.8 & 78.82 & 83.76 & 73.13 & 77.15 \\
        AdvLoRA+Att & 2.8M & 66.2 & 82.5 & 85.8 & 78.17 & 57.7 & 77.52 & 83.32 & 72.85 & 75.51 \\
        AdvLoRA*+Att & 2.8M & \underline{71} & \underline{86.8} & \textbf{90.7} & \textbf{82.83} & 58.02 & \underline{80.1} & \textbf{85.9} & \underline{74.67} & \textbf{78.75} \\ \hline
    \end{tabular}
    }
    \label{advlora_flickr_att}
\end{table*}

\subsection{Performance Comparisons}

We compare our method with five baselines across two cross-modal retrieval tasks using four datasets. Specifically, we perform adversarial adaptation based on the PGD-3 attack for all methods and evaluate their performance under adversarial conditions.

\textbf{Evaluation on Image-Text Retrieval.} We conducted experiments on adversarially attacked data for both MSCOCO and Flickr30K, as presented in Tab.\ref{advlora_coco_att} and Tab.\ref{advlora_flickr_att}. The results lead us to two key conclusions: (1) \textit{Performance Under Adversarial Attacks.} After adaptation, AdvLoRA outperforms all other baselines when confronted with adversarial attacks. On MSCOCO, AdvLoRA surpasses other PEFT methods by 12.17\% and exceeds FFT by 2.47\%, using approximately 100 times fewer tunable parameters. (2) \textit{Enhanced Robustness with Larger Datasets.} AdvLoRA shows improved robustness on larger datasets, highlighting the potential of PEFT methods to bolster model resilience. On the smaller Flickr30K dataset, the performance of various baselines post-adaptation remains comparable, with no significant robustness increase. However, on MSCOCO, FFT achieves notable robustness but still trails behind AdvLoRA. These results emphasize AdvLoRA's advantages in clustering reparameterization and parameter alignment.

\textbf{Evaluation on Video-Text Retrieval.} We conducted experiments on adversarially attacked data for both DiDeMo and MSR-VTT, as presented in Tab.\ref{advlora_msrvtt_att} and Tab.\ref{advlora_didemo_att}. Our findings lead to the following conclusions: (1) \textit{Adversarial Robustness of AdvLoRA.} AdvLoRA demonstrates excellent adversarial robustness on video data, surpassing all other baselines. In DiDeMo, AdvLoRA slightly outperforms UniAdapter while utilizing seven times fewer parameters. On MSR-VTT, AdvLoRA enhances the model's adversarial robustness by 39.16\%, significantly exceeding other baselines. (2) \textit{Improved Robustness with Larger Datasets.} AdvLoRA shows better adversarial robustness on larger datasets. On the smaller DiDeMo dataset, the performance of various baselines after adversarial adaptation remains comparable, with minimal improvements in robustness. In contrast, on the larger MSR-VTT dataset, while UniAdapter achieves notable adversarial robustness, it still falls short of AdvLoRA's performance, using seven times more parameters. These results can be attributed to AdvLoRA's design, particularly in terms of clustering reparameterization and parameter alignment. These results further highlight the design advantages of AdvLoRA in clustering reparameterization and parameter alignment.

\begin{table}[h]
\caption{Ablation study on AdvLoRA components. PC, PA, and PU denote parameter clustering, parameter alignment, and adaptive parameter update, respectively. The metrics TR@Mean and VR@Mean denote the mean recall for text-to-video and video-to-text tasks, respectively, while R@Mean represents the average of TR@Mean and VR@Mean.}
\scalebox{0.85}{
  \begin{tabular}{c|c|c|c|c|c|c}
  \hline
    Setting & PC & PA & PU & TR@Mean & VR@Mean & R@Mean \\
    \midrule
    Baseline & \ding{55} & \ding{55} & \ding{55} & 54.37 & 54.07 & 54.22 \\
    \hline
    a & \ding{51} & \ding{55} & \ding{55} & 55.36 & 56.29 & 55.83 \\
    b & \ding{51} & \ding{51} & \ding{55} & \textbf{57.85} & 57.21 & 57.53\\
    c & \ding{51} & \ding{51} & \ding{51} & 57.69 & \textbf{58.42} & \textbf{58.01} \\
  \bottomrule
  \end{tabular}
}
\vspace{-1.0em}
\label{ablation}
\end{table}

\vspace{-0.5em}
\subsection{Ablation Study}

\textbf{Effect of Proposed Module.}
To demonstrate the effectiveness of the proposed module, we conduct an ablation study on the components of AdvLoRA using the Didemo dataset and summarize our results in Tab.~\ref{ablation}. Firstly, we explored the impact of the parameter clustering technique. Our experiments, comparing model variant (a) with the baseline, indicate a notable improvement when leveraging knowledge from pre-trained models through parameter clustering. Moreover, the performance gains observed in model variants (a), (b), and (c) highlight that incorporating parameter alignment and adaptive parameter update methods on top of parameter clustering significantly enhances the adversarial adaptability of VLMs. Finally, using all the proposed methods together, i.e. AdvLoRA, results in the best overall performance, achieving a mean recall of 58.01.

\begin{figure}[h]
\centering
\begin{minipage}{0.49\linewidth}
    \centering
    \includegraphics[width=0.9\linewidth]{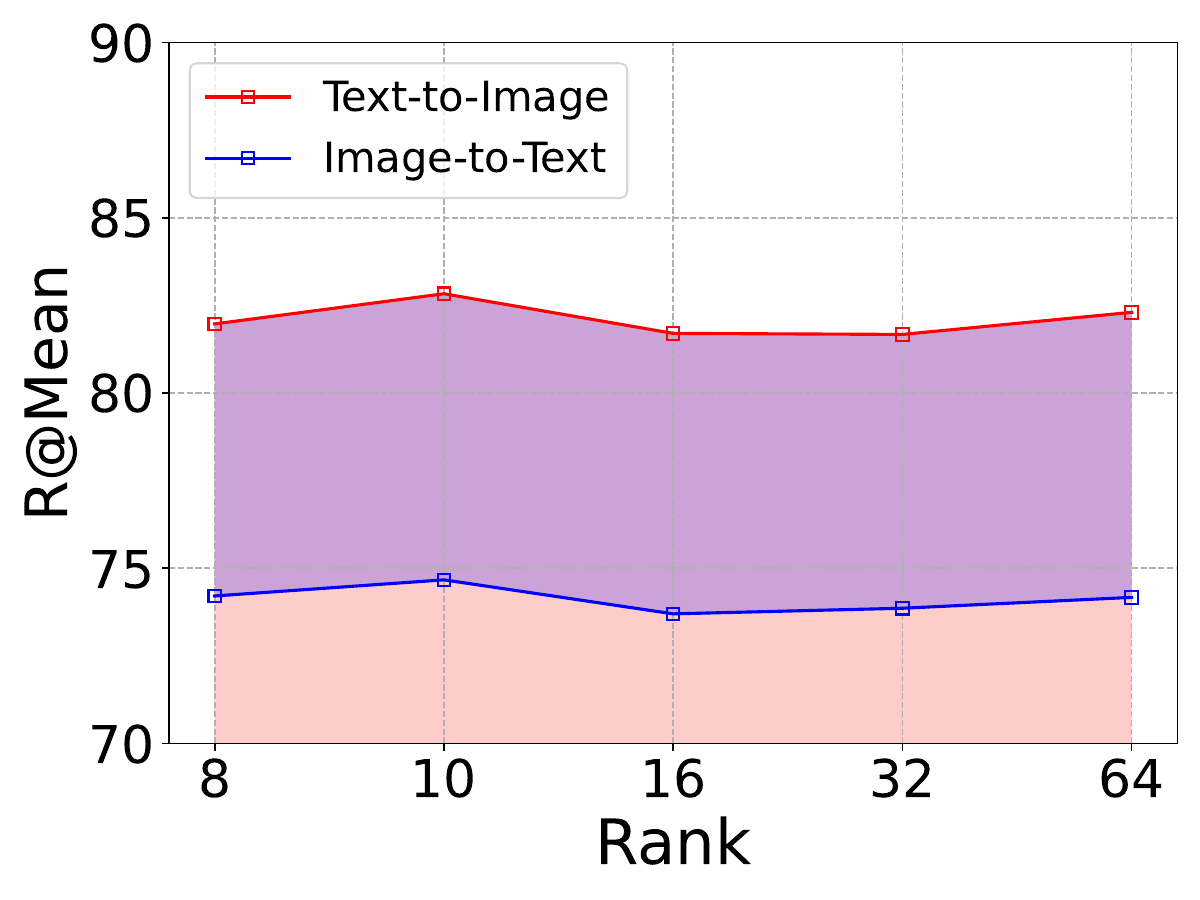}
    \caption{Sensitivity Analysis.}
    \label{fig:rank}
\end{minipage}
%\qquad
\begin{minipage}{0.49\linewidth}
    \centering
    \includegraphics[width=0.9\linewidth]{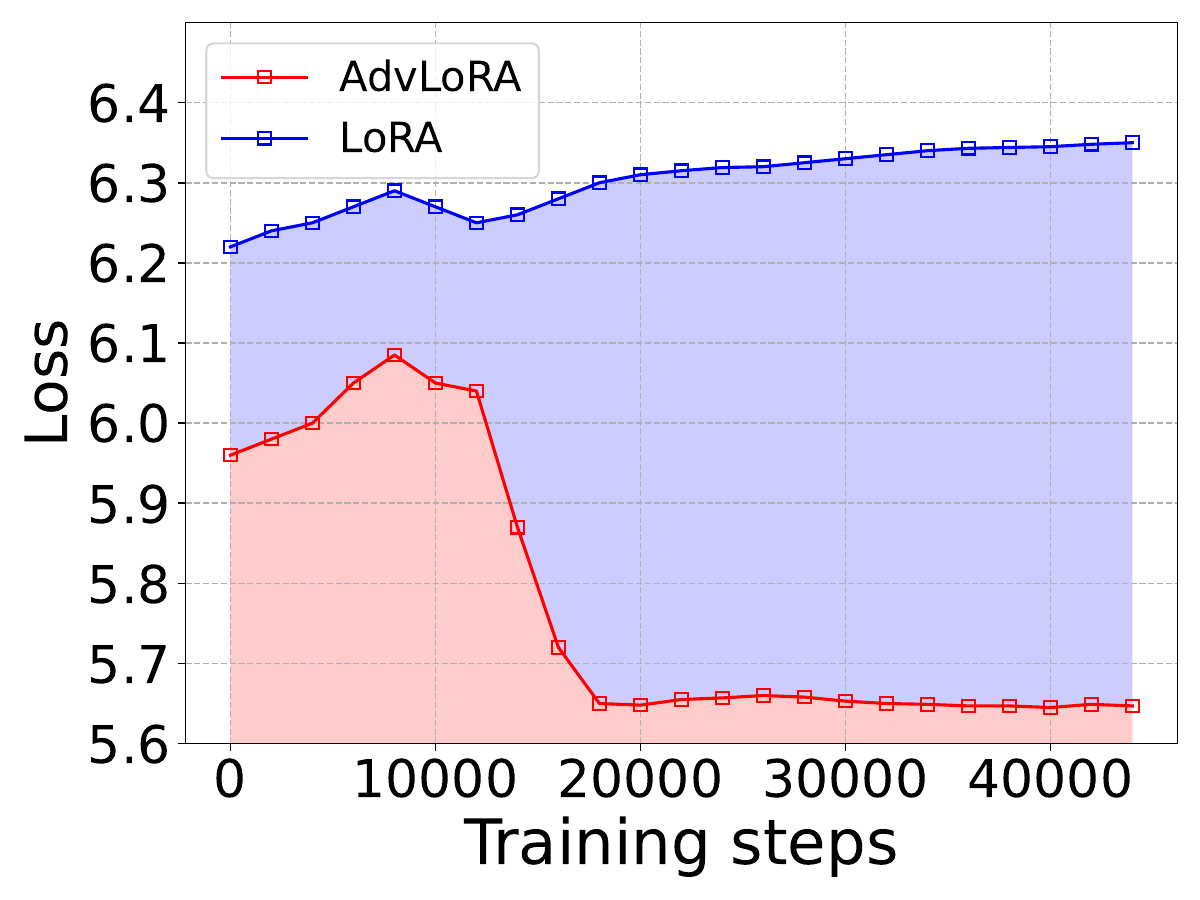}
    \caption{Loss Analysis.}
    \label{fig:loss}
\end{minipage}
\vspace{-1.8em}
\end{figure}

\textbf{Sensitivity Analysis of Rank Size.}
To demonstrate the impact of different rank sizes on AdvLoRA, we trained a series of models with varying ranks on Flickr30K. The results are shown in Fig.~\ref{fig:rank}. We find that as the rank size increases, the number of tunable parameters increases, but AdvLoRA's performance remains stable, while computational cost rises. Therefore, during adversarial adaptation, performance is not sensitive to rank size, allowing us to choose a smaller rank without significant loss in performance.

\textbf{Loss Convergence Analysis.}
To explore the effectiveness of the proposed method in terms of loss convergence, we examine two model variants, LoRA and AdvLoRA, on the Flickr30K dataset. The results are presented in Fig.~\ref{fig:loss}. The experimental findings indicate that (1) AdvLoRA achieves a lower loss early in the training process, demonstrating that techniques we proposed such as parameter clustering can serve as an effective "warm-up," helping VLMs establish a favorable starting point during adversarial adaptation; (2) as training progresses, AdvLoRA successfully navigates to a more effective gradient descent trajectory, leading to more efficient model training. In contrast, LoRA appears to get trapped in a local optimum. This highlights the superiority of our proposed method in facilitating better convergence of VLMs.

\begin{table}[t]
    \caption{Comparison on the training time and GPU memory.}
    \centering
    \scalebox{0.88}{
    \begin{tabular}{lccc}
    \hline
        Method & Tuable Para. & Time & Mem. \\ \hline
        BLIP+FFT & 223M & 1.00 & 1.00 \\ 
        BLIP+LoRA & 2.8M & 0.91 & 0.85 \\ 
        BLIP+LP & 0.5M & 0.79 & 0.67 \\ 
        Uniadapter & 19.5M & 0.93 & 0.77 \\ 
        Aurora & 0.3M & 1.05 & 1.04 \\ 
        AdvLoRA & 2.8M & 0.94 & 0.85 \\ \hline
    \end{tabular}
    }
    \label{gpu}
    \vspace{-1.5em}
\end{table}

\begin{table*}[t]
    \caption{Adversarial experiment on MSR-VTT. An asterisk (*) indicates that adversarial adaptation has been performed. The best results are displayed in bold, while the second-best results are underlined.}
    \centering
    \scalebox{0.88}{
    \begin{tabular}{lcccccccccc}
    \hline
        \multirow{2}{*}{Method} & \multirow{2}{*}{Tunable Para.} & \multicolumn{4}{c}{MSR-VTT TR} & \multicolumn{4}{c}{MSR-VTT VR} \\
        ~ & ~ & R@1 & R@5 & R@10 & R@Mean & R@1 & R@5 & R@10 & R@Mean & Mean \\ \hline
        BLIP+FFT+Att & 223M & 1.2 & 5 & 7.6 & 4.6 & 2.7 & 8.1 & 12.5 & 7.77 & 6.18 \\
        BLIP+FFT*+Att & 223M & 21 & 41.9 & 50.8 & 37.9 & 21 & 46.8 & 57.9 & 41.9 & 39.9 \\
        BLIP+LoRA+Att & 2.8M & 12.8 & 23.4 & 28.1 & 21.43 & 18.9 & 30.8 & 37.8 & 29.16 & 25.30 \\
        BLIP+LoRA*+Att & 2.8M & 21.2 & 43.5 & 52.7 & 39.13 & 21 & 42.5 & 52.1 & 38.53 & 38.83 \\
        BLIP+LP+Att & 0.5M & 7.7 & 16.1 & 20.1 & 14.63 & 14.4 & 26.4 & 32.8 & 24.53 & 19.58 \\
        BLIP+LP*+Att & 0.5M & 14.5 & 26.8 & 33.3 & 24.87 & 15.8 & 26.7 & 33.5 & 25.33 & 25.10 \\
        UniAdapter+Att & 19.5M & 8.3 & 15.4 & 18.9 & 14.2 & 11.6 & 22.6 & 27.2 & 20.47 & 17.33 \\
        UniAdapter*+Att & 19.5M & \underline{38.6} & \underline{64} & \underline{74.5} & \underline{59.03} & \underline{39.2} & \underline{64.9} & \underline{75.8} & \underline{59.97} & \underline{59.50} \\
        Aurora+Att & 0.6M & 11.6 & 20.3 & 24.6 & 18.83 & 16.9 & 30.1 & 36.7 & 27.9 & 23.37 \\
        Aurora*+Att & 0.6M & 38.1 & 63.6 & 73.5 & 58.4 & 37 & 60.8 & 72.7 & 56.83 & 57.62 \\
        AdvLoRA+Att & 2.8M & 12.3 & 21.8 & 26.2 & 20.1 & 15.8 & 28.4 & 34.2 & 26.13 & 23.12 \\
        AdvLoRA*+Att & 2.8M & \textbf{40.4} & \textbf{67.4} & \textbf{78.6} & \textbf{62.13} & \textbf{40.5} & \textbf{68.4} & \textbf{78.4} & \textbf{62.43} & \textbf{62.28} \\ \hline
    \end{tabular}
    }
    \label{advlora_msrvtt_att}
\vspace{-0.5em}
\end{table*}

\begin{table*}[t]
    \caption{Adversarial experiment on Didemo. An asterisk (*) indicates that adversarial adaptation has been performed. The best results are displayed in bold, while the second-best results are underlined.}
    \centering
    \scalebox{0.88}{
    \begin{tabular}{lcccccccccc}
    \hline
        \multirow{2}{*}{Method} & \multirow{2}{*}{Tunable Para.} & \multicolumn{4}{c}{Didemo TR} & \multicolumn{4}{c}{Didemo VR} \\
        ~ & ~ & R@1 & R@5 & R@10 & R@Mean & R@1 & R@5 & R@10 & R@Mean & Mean \\ \hline
        BLIP+FFT+Att & 223M & 12.66 & 26.32 & 35.39 & 24.79 & 14.56 & 31.7 & 40.58 & 28.95 & 26.87 \\
        BLIP+FFT*+Att & 223M & 29.71 & 53.04 & 64.3 & 49.08 & 31.21 & 55.63 & 67.4 & 51.41 & 50.25 \\
        BLIP+LoRA+Att & 2.8M & 33.2 & 57.43 & 66.7 & 52.44 & 32.7 & 56.73 & 68.1 & 52.51 & 52.48 \\
        BLIP+LoRA*+Att & 2.8M & 33.7 & 59.82 & 69.59 & 54.37 & 32.8 & 59.02 & 70.39 & 54.07 & 54.22 \\
        BLIP+LP+Att & 0.5M & 23.13 & 45.86 & 53.54 & 40.84 & 26.02 & 47.06 & 57.03 & 43.37 & 42.11 \\
        BLIP+LP*+Att & 0.5M & 22.73 & 45.46 & 54.04 & 40.74 & 25.32 & 46.46 & 56.73 & 42.84 & 41.79 \\
        UniAdapter+Att & 19.5M & 27.02 & 52.14 & 64.01 & 47.72 & 9.27 & 24.83 & 36.69 & 23.6 & 35.66 \\
        UniAdapter*+Att & 19.5M & \underline{36.38} & \underline{63.5} & \textbf{73.57} & \underline{57.82} & {35.88} & \textbf{64.3} & \textbf{73.87} & \textbf{58.02} & \underline{57.92} \\
        Aurora+Att & 0.6M & 30.31 & 52.94 & 64.11 & 49.12 & 31.21 & 54.74 & 64.61 & 50.19 & 49.65 \\
        Aurora*+Att & 0.6M & 35.59 & 61.22 & 72.18 & 56.33 & \underline{36.69} & 62.01 & 71.88 & 56.86 & 56.60 \\
        AdvLoRA+Att & 2.8M & 34.4 & 62.11 & 71.39 & 55.97 & 35.19 & 62.81 & 70.99 & 56.33 & 56.15 \\
        AdvLoRA*+Att & 2.8M & \textbf{37.38} & \textbf{64.4} & \underline{73.48} & \textbf{58.42} & \textbf{36.99} & \underline{63.21} & \underline{72.88} & \underline{57.69} & \textbf{58.06} \\ \hline
    \end{tabular}
    }
    \label{advlora_didemo_att}
\vspace{-0.5em}
\end{table*}

\textbf{Efficiency and Storage Cost Analysis.}
To investigate the cost-effectiveness of the proposed method, we analyze the relative training GPU hours and GPU memory costs of AdvLoRA compared to five baseline models on Flickr30K. The results are presented in Tab.~\ref{gpu}, where the time (or memory) of FFT is normalized to one unit. The findings indicate that: (1) AdvLoRA exhibits relatively low parameter count and memory overhead; (2) AdvLoRA incurs longer time costs due to the necessity of offline clustering reparameterization and parameter alignment prior to adaptation.

\begin{table}[ht]
\caption{Additional attack types on the MSCOCO dataset. ``TR'' and ``IR'' donate text-to-image retrieval and image-to-text retrieval.}
    \centering
    \renewcommand{\arraystretch}{1.3}
    \scalebox{0.88}{
    \begin{tabular}{llccc}
    \hline
        Method & Attack & TR@Mean & IR@Mean & Mean \\ \hline
        
        LoRA    & PGD-3 &   61.01 &  54.41 & 57.71 \\
        AdvLoRA & PGD-3  & {82.40}  & {69.83} & {76.12} \\ \hline
        
        AdvLoRA & PGD-20  & {81.65}  & {69.13} & {75.39} \\ 
        AdvLoRA & FGSM  & {84.44}  & {72.21} & {78.32} \\ 
        AdvLoRA & BIM  & {83.25} & {69.56} & {76.41} \\ \hline
        
        AdvLoRA & SA  & {87.40} &  {75.83} & {81.62} \\ 
        AdvLoRA & ZOO  &  {84.17} & {73.83} & {79.00} \\ \hline

    \end{tabular}
    }
    \label{buchong}
\vspace{-0.5em}
\end{table}

\label{more_attack}
\textbf{Performance on More Attacks.}
To demonstrate the robust generalization of the proposed method, we apply additional attacks to the AdvLoRA model, which has undergone adversarial adaptation on the MSCOCO dataset. These include white-box attack methods (FSGM~\cite{goodfellow2014explaining}, PGD-20~\cite{madry2017towards}, BIM~\cite{kurakin2016adversarial}) and black-box methods (Zeroth Order Optimization (ZOO)~\cite{chen2017zoo}, SquareAttack with 500 queries (SA)~\cite{andriushchenko2020square}), as shown in Tab.~\ref{buchong}. Note that AdvLoRA is adapted under PGD-3. The experimental results indicate that (1) AdvLoRA consistently outperforms the baseline model, LoRA, across various attack types. This trend continues with the FGSM attack, where AdvLoRA scores 84.44 and 72.21, compared to LoRA's 61.01 and 54.41. (2) AdvLoRA also demonstrates strong performance under black-box attacks. For the SA, AdvLoRA attains a TR mean score of 87.40 and an IR mean score of 75.83, highlighting its robustness against unseen attack strategies. The results collectively affirm that AdvLoRA not only enhances robustness through adversarial adaptation but also generalizes effectively across different attack types, thereby reinforcing the model’s applicability in real-world scenarios where diverse adversarial threats may be encountered.

\label{larger_backbones}
\textbf{Scaling to Larger Backbones.}
To demonstrate the scalability of the proposed method, we evaluated various sizes of BLIP as the backbone on Flickr30K, as shown in Tab.~\ref{backbones}. The findings indicate that: (1) \textit{Consistent Performance Improvement}. AdvLoRA consistently outperforms the baseline model, LoRA, across both TR and IR metrics, achieving notable scores for different backbone sizes. (2) \textit{Enhanced Capability Utilization.} This trend continues with the larger BLIP architectures, where AdvLoRA maintains a clear performance edge over LoRA. (3) \textit{Significant Gains in Larger Architectures.} When scaling to the larger BLIP-2-opt-2.7b backbone, AdvLoRA shows substantial improvements, reaffirming its effectiveness. These results highlight AdvLoRA's robustness and adaptability, confirming its ability to leverage larger architectures for enhanced retrieval performance. This scalability not only improves model efficiency but also supports its application in complex real-world scenarios, where capturing intricate data patterns is crucial.

\begin{table}[t]
\caption{Performance on Flickr30K when scaling to larger backbone networks. ``TR'' and ``IR'' donate text-to-image retrieval and image-to-text retrieval.}
    \centering
    \renewcommand{\arraystretch}{1.3}
    \scalebox{0.85}{
    \begin{tabular}{llccc}
    \hline
        Backbone & Method & TR@Mean & IR@Mean & Mean \\ \hline

        \multirow{2}{*}{BLIP-base} & LoRA &  80.40 & 73.24 & 76.82 \\
        ~ & AdvLoRA &  82.83 & 74.67 & 78.75 \\ \hline

        \multirow{2}{*}{BLIP-large} & LoRA &  81.73 & 73.01 & 77.34 \\
        ~ & AdvLoRA & 84.74  & 75.56 & 80.15 \\ \hline

        \multirow{2}{*}{BLIP-2-opt-2.7b} & LoRA & 83.32 & 81.18 & 82.25 \\
        ~ & AdvLoRA &  87.34 & 83.66 & 85.50 \\ \hline

    \end{tabular}
    }
    \label{backbones}
\end{table}

\textbf{Performance on the Natural Data.} 
To evaluate the performance of our proposed method on natural images, we assess AdvLoRA and five baselines on MSCOCO, as shown in Tab.~\ref{coco_nat}. The findings indicate: (1) \textit{Trade-off in Performance.} While adversarial adaptation enhances robustness against specific attacks, it can degrade performance on natural data, suggesting a trade-off between adversarial resilience and generalization. (2) \textit{Superior Metrics of AdvLoRA.} AdvLoRA achieves a TR mean score of 85.18 and an IR mean score of 79.92, surpassing all baselines due to its ability to learn semantically invariant feature representations. (3) \textit{Versatility Across Retrieval Tasks.} Compared to methods like BLIP+FFT and UniAdapter, AdvLoRA excels in TR and IR metrics, demonstrating a retrieval mean score of 79.92, underscoring its reliability. These results affirm AdvLoRA's effectiveness in balancing robustness and performance in natural image retrieval.

\begin{table}[t]
\caption{Performance on Flickr30K when scaling to larger backbone networks. ``TR'' and ``IR'' donate text-to-image retrieval and image-to-text retrieval.}
    \centering
    \renewcommand{\arraystretch}{1.3}
    \scalebox{0.88}{
    \begin{tabular}{lccc}
    \hline
        Method & TR@Mean & IR@Mean & R@Mean \\ \hline
        
        BLIP+FFT*+Nat. & 74.17  & 69.11 & 71.65 \\
        BLIP+LoRA*+Nat. & 85.29 & 74.75 & 80.02 \\
        BLIP+LP*+Nat. & 86.33 & 75.38 & 80.86 \\
        UniAdapter*+Nat. & 75.29 & 66.89 & 71.09 \\
        Aurora*+Nat. & 84.74 & 73.54 & 79.15 \\
        AdvLoRA*+Nat. & 85.18 & 74.67 & 79.92 \\ \hline
    \end{tabular}
    }
    \label{coco_nat}
%    \vspace{-1em}
\end{table}

\begin{figure}[t]
        \centering
        \includegraphics[width=0.97\linewidth]{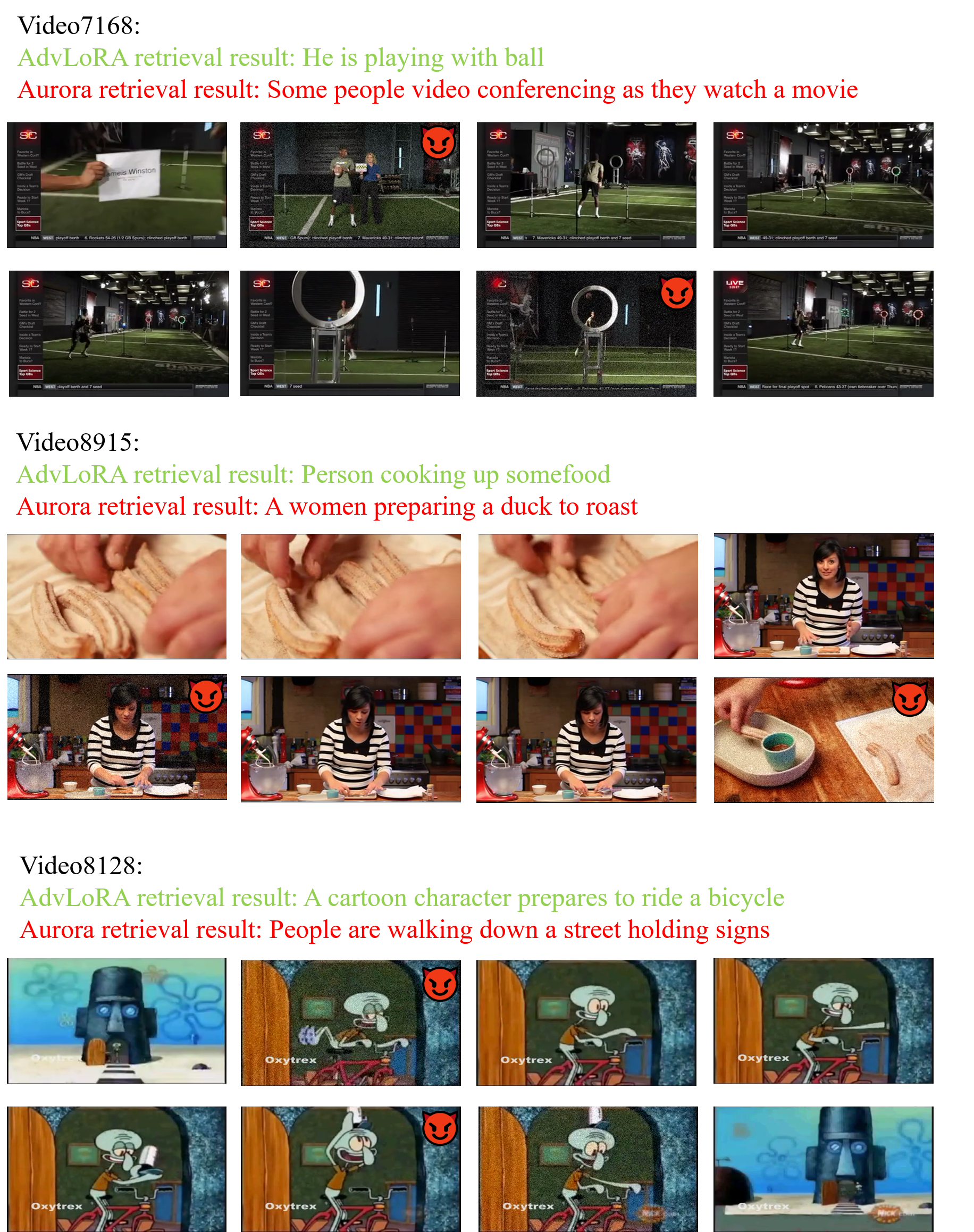}
        \caption{Case study of MSR-VTT. We sample and visualize eight frames from the videos. The frames with the devil denote that they are under the adversarial attacks.}
        \label{case}
        \vspace{-1em}
\end{figure}

\textbf{Case Study.}
To visually demonstrate the effectiveness of AdvLoRA, we present several test cases from the MSR-VTT dataset, as illustrated in Fig.~\ref{case}. The results show that: (1) \textit{Resilience to Adversarial Attacks.} In Video168, AdvLoRA retrieves "He is playing with ball," accurately reflecting the scene despite adversarial perturbations, while Aurora shows a noticeable decline in performance. (2) \textit{Contextual Understanding.} For Video8915, AdvLoRA identifies "Women preparing to cook a roast," demonstrating its ability to grasp contextual details, whereas Aurora's output lacks the same level of accuracy. (3) \textit{Complex Scene Interpretation.} In Video128, AdvLoRA maintains clarity in retrieving a cartoon character preparing to ride a bicycle, contrasting with Aurora’s irrelevant result, highlighting AdvLoRA’s ability to retain semantic integrity amid adversarial challenges. These examples underscore AdvLoRA's robustness and contextual awareness in dynamic environments.

\section{Conclusion}
In this paper, we aim to alleviate the security risks in the Vision-Language Models (VLMs). First of all, we show the vulnerability of VLMs with various adaptation methods under adversarial attacks via extensive experiments. Besides, as the sizes of VLMs increase, simply applying the conventional adversarial adaptation methods to VLMs easily leads to (1) unpromising adversarial robustness and (2) tremendous parameter costs. To tackle these issues, we propose a novel parameter-efficient adversarial adaptation method called AdvLoRA, which incorporates parameter clustering, parameter alignment, and adaptive parameter updates. Extensive experiments validate the effectiveness and efficiency of AdvLoRA, revealing the intrinsic low-rank property that emerges during the adversarial adaptation process. Our technique, which emphasizes clustering reparameterization and parameter alignment, significantly enhances the adaptation process. This work provides a fresh perspective for researchers focused on security within the broader context of Artificial General Intelligence (AGI). Moving forward, we aim to optimize memory and computational resources during the adaptation process and enhance the adversarial robustness of VLMs, particularly against language-based attacks.

\begin{acks}
This work is supported by the Natural Science Foundation of China under Grant Nos. 72225011, 72434005 and L242400108.
\end{acks}
\bibliographystyle{ACM-Reference-Format}
\bibliography{acmart}

\end{document}